\documentclass[final,3p,times,twocolumn]{elsarticle} 
\usepackage{booktabs}           
\usepackage{array}

\usepackage{amssymb}
\usepackage[T1]{fontenc}        
\usepackage{amsmath}
\usepackage{booktabs}
\usepackage{tabularray}
\usepackage{amssymb}
\usepackage{amsmath}
\usepackage{subcaption}
\usepackage{graphicx}
\usepackage{multirow}           
\usepackage{multicol}
\usepackage{url}                
\usepackage{hyperref}           

\usepackage{breakurl}
\usepackage{comment}
\usepackage[none]{hyphenat}
\usepackage{adjustbox}
\usepackage{microtype}
\usepackage{ragged2e}
\usepackage{xcolor}

\journal{arXiv}

\begin{document}

\begin{frontmatter}

\title{MAD-SmaAt-GNet: A Multimodal Advection-Guided Neural Network for Precipitation Nowcasting}


\author{Samuel van Wonderen}
\ead{samuvw@outlook.com}
\author{Siamak Mehrkanoon\corref{cor1}}
\ead{s.mehrkanoon@uu.nl}
\cortext[cor1]{Corresponding author}

\address{Department of Information and Computing Sciences, Utrecht University, Utrecht, The Netherlands}

\begin{abstract}
Precipitation nowcasting (short-term forecasting) is still often performed using numerical solvers for physical equations, which are computationally expensive and make limited use of the large volumes of available weather data. Deep learning models have shown strong potential for precipitation nowcasting, offering both accuracy and computational efficiency. Among these models, convolutional neural networks (CNNs) are particularly effective for image-to-image prediction tasks. The SmaAt-UNet is a lightweight CNN based architecture that has demonstrated strong performance for precipitation nowcasting. This paper introduces the Multimodal Advection-Guided Small Attention GNet (MAD-SmaAt-GNet), which extends the core SmaAt-UNet by (i) incorporating an additional encoder to learn from multiple weather variables and (ii) integrating a physics-based advection component to ensure physically consistent predictions. We show that each extension individually improves rainfall forecasts and that their combination yields further gains. MAD-SmaAt-GNet reduces the mean squared error (MSE) by 8.9\% compared with the baseline SmaAt-UNet for four-step precipitation forecasting up to four hours ahead. Additionally, experiments indicate that multimodal inputs are particularly beneficial for short lead times, while the advection-based component enhances performance across both short and long forecasting horizons.

   
\end{abstract}


\end{frontmatter}

\section{Introduction} \label{sec:Intro} 

In present-day weather forecasting, complex and computationally intensive models are used that rely on physics-based equations describing relationships among multiple atmospheric variables. These equations are solved numerically given known or estimated initial and boundary conditions \cite{atmos15060689, doi:10.1126/science.adi2336, jones2017machine, sun2014useNWP}. However, numerical solvers require substantial time and computational resources, which limits their suitability for very short-term forecasting, or nowcasting \cite{Espeholt2022, ravuri2021skilful, Zhang2023526, atmos15060689, severijns2023short, sun2014useNWP}. For this reason, among others, researchers have increasingly turned to deep learning to enable rapid short-horizon predictions \cite{An2025SurveyPrecip, Espeholt2022, Zhang2023526, Bi2023533, doi:10.1126/science.adi2336, NIU2025M4Caster}. The present study contributes to this line of work by introducing a model that incorporates physical knowledge in one of its components and leverages additional weather variables as multimodal inputs, both of which are shown to improve precipitation forecasts.

Deep learning models offer major advantages for weather prediction: once trained, they can generate new forecasts within seconds \cite{Zhang2023526, atmos15060689, Espeholt2022, Bi2023533}, and they can learn atmospheric dynamics directly from data, including relationships not explicitly formulated in Numerical Weather Prediction (NWP) models \cite{Zhang2023526, atmos15060689, Espeholt2022, ruiwangPhysicsGuided, Bi2023533, doi:10.1126/science.adi2336}. However, purely data-driven approaches do not explicitly encode the domain knowledge captured by physical equations. Such knowledge can constrain predictions to physically plausible states \cite{ruiwangPhysicsGuided, Zhang2023526, Tanyu_2023, 10.1145/3394486.3403198}, improve forecast accuracy, and reduce the effective search space of model parameters \cite{ruiwangPhysicsGuided}.

These considerations have motivated the emergence of Physics-Informed (or Physics-Guided) Deep Learning \cite{Bézenac2019, 10.1145/3394486.3403198}, and within it the weather-specific field of Machine/Deep Learning Weather Prediction \cite{doi:10.1126/science.adi2336}. In this area, models integrate data-driven learning with physical knowledge in the form of differential equations, physical constraints, symmetries, or known monotonic relations among variables \cite{Tanyu_2023, Panghal20212989}.

Precipitation forecasting, particularly quantitative precipitation nowcasting (QPN), is a central task in this field due to the societal and operational importance of predicting rain, snow, and hail \cite{Li20214019, atmos15060689, FERNANDEZ2021BroadUnet, doi:10.1126/science.adi2336, niu2023heterogeneous, jones2017machine}. The task is challenging because precipitation systems exhibit nonlinear behavior across multiple spatial and temporal scales \cite{NIU2025M4Caster}. Radar imagery is widely used in QPN \cite{doi:10.1126/science.adi2336} because it is rapidly available and provides a reliable proxy for precipitation intensity \cite{NIU2025M4Caster, Li20214019, TrebingSmaAtUNet, Klein_2015_CVPR, NIU2025M4Caster, zhao2024new, niu2023heterogeneous, ravuri2021skilful, Zhang2023526, xu2022radar}.

Researchers have explored different strategies for incorporating physical knowledge into neural networks for precipitation nowcasting. Some models impose physical constraints such as the 2D continuity equation or infer motion fields (optical flow) and use them to advect radar images \cite{Bézenac2019, Li20214019}. The NowcastNet model, for example, includes an evolution network that enforces the 2D continuity equation while learning motion fields and intensity residuals to generate future radar frames \cite{Zhang2023526}.

Another major direction is multimodal fusion, where multiple data sources are combined to enrich the predictions \cite{Li20214019, NIU2025M4Caster, zhao2024new, Kaparakis2023233, NIU2025M4Caster, niu2023heterogeneous}. For instance, \cite{Li20214019} and \cite{NIU2025M4Caster} combine radar and satellite data to produce more accurate precipitation maps, while other studies include wind fields \cite{Kaparakis2023233} or precipitation masks \cite{REULEN2024112612}. This suggests that additional weather variables can further enhance precipitation nowcasting performance.

These two directions remain major frontiers in the QPN field, with ongoing research continually introducing refinements to model architectures and data-fusion strategies \cite{NIU2025M4Caster, niu2023heterogeneous, NIU2025M4Caster, zhao2024new, Zhang2023526, niu2024fsrgan, Ma2023MM_RNN}. However, to the best of our knowledge, a single model that jointly integrates physics-informed components and multimodal weather inputs has not yet been fully explored. This study investigates such a combined approach by incorporating the physically consistent evolution network of NowcastNet \cite{Zhang2023526} into a model that also fuses rain images with additional weather variables, namely temperature, air pressure, wind speed, and relative humidity. Our experiments show that including these additional variables improves forecasting performance and that the combined physics-informed and multimodal model outperforms models using only one of the two strategies.

This paper is structured as follows. Section \ref{sec:RelWork} reviews related research works on data driven weather forecasting using machine learning models. Section \ref{sec:Methods} details the model architecture, the ablation study, the training procedure, and provides an overview of model parameter counts and evaluation metrics. Section \ref{sec:Exper} describes the dataset and the prediction task. Section \ref{sec:ResuDiscu} presents and discusses the experimental results. Finally, Section \ref{sec:Conclu} summarises the conclusions and offers recommendations for future work.

\section{Related works} \label{sec:RelWork}
A few early works on implementing CNNs for weather prediction are described in \cite{Klein_2015_CVPR, MEHRKANOON2019120} where, in the former, a CNN with dynamic filters is used to predict a single target precipitation image and, in the latter, a 1D, 2D, and 3D CNN model are developed to predict future temperatures and wind speeds at given weather stations from historical data of the aforementioned and other weather variables.

The paper \cite{TrebingWind} continues from the work of \cite{MEHRKANOON2019120} to create a CNN model that implements Depthwise Separable Convolutions \cite{CholletDSC} (DSCs) which first apply convolutions over the height and width of a tensor, per channel, and then point-wise convolutions across channels, thereby capturing spatial and cross-channel relations. Here, the input is a number of weather variables per cities and per time steps and the target is wind speed predictions for three cities at a specified forecast time. The model was compared to several CNN models and it was shown that the model performed best for higher forecast times and on average while having a comparable or smaller number of parameters.

The authors of \cite{TrebingSmaAtUNet} introduce SmaAt-UNet, which is used as the core model in our study, and has since served as the basis for many subsequent models \cite{FERNANDEZ2021BroadUnet, REULEN2024112612, turzi2025SSA_UNet, Yimin2022TransUNet}. Motivated by the success in image-related tasks and even in precipitation nowcasting \cite{agrawal2019machine}, the authors in \cite{TrebingSmaAtUNet} developed a U-Net \cite{RonnebergerUnet} based model for precipitation nowcasting but enhanced it with more parameter efficiency by implementing DSCs \cite{CholletDSC} and Convolutional Block Attention Modules \cite{WooCBAM} (CBAMs). In fact, the model performs on par or better than the standard U-Net model in the nowcasting task of a single rain image with only a fourth of the number of parameters. 

The encoder of a standard U-Net \cite{RonnebergerUnet} consists of four downsampling operations that halve the spatial dimensions of the input, followed by a double convolutional layer. The decoder has four upsampling operations that apply bilinear interpolation to double the spatial dimensions of the input followed by a similar double convolution layer. In contrast to the standard U-Net, the SmaAt-UNet uses DSCs \cite{CholletDSC} in its double convolutions and there is a CBAM \cite{WooCBAM} after each downsampling operation that is applied to the output. A CBAM \cite{WooCBAM} enables the model to identify relevant channels to the output via channel-attention in the form of a vector that gives more weight to certain channels of the input. Additionally, a CBAM \cite{WooCBAM} subsequently applies spatial attention by multiplying the input images with a matrix to identify spatial regions relevant to the output. As such, the decoder receives inputs from the skip-connections and the bottleneck that have been treated with channel-attention and spatial-attention, allowing for richer or less noisy features.

In \cite{Kaparakis2023233}, the authors introduce a 3D-CNN based U-Net architecture that combines radar-derived precipitation fields with wind-speed data to predict future precipitation at a specified lead time. The model processes precipitation and wind speed data through two separate U-Net branches, whose outputs are then fused and mapped to a single target precipitation image. The proposed architecture is evaluated against two benchmark CNN models and against its own U-Net variant without wind-speed input, and it achieves the best performance, though its accuracy decreases as the prediction horizon increases.

The authors of \cite{REULEN2024112612} extend the SmaAt-UNet architecture of \cite{TrebingSmaAtUNet} by incorporating binary precipitation masks as additional inputs, inspired in part by the multimodal design in \cite{Kaparakis2023233}. Their model, termed SmaAt-GNet, adds a second encoder dedicated to the rain masks and employs a deeper decoder than the original SmaAt-UNet. They further integrate this architecture into a conditional Generative Adversarial Network \cite{GoodfellowGAN} (cGAN), forming their full model, GA-SmaAt-GNet.

Similar to SmaAt-UNet, the GA-SmaAt-GNet takes sequences of rain images as input, but it also generates corresponding rain masks. These masks are produced by accumulating the input frames into a single composite image and applying multiple rain-rate thresholds to obtain binary masks at different intensity levels. Rain images and masks are processed by their respective encoders, and their feature outputs are concatenated at the bottleneck and at each skip connection before being passed to the decoder. The decoder predicts future rain images, which are evaluated by a discriminator trained to distinguish real from generated sequences. The generator (SmaAt-GNet) simultaneously learns to fool the discriminator while minimising the MSE of its predictions, resulting in outputs that are both accurate and visually realistic. The model is compared to several benchmark models, including SmaAt-UNet and its SmaAt-GNet without the discriminator. It is shown that the model performs best in almost all cases, especially the samples with extreme precipitation.



In \cite{Zhang2023526}, NowcastNet is introduced as a physics-informed cGAN \cite{GoodfellowGAN} model for extreme precipitation nowcasting. The evolution network component of this model is used for our study as a physics-based component to our model. The evolution network produces physically consistent future rain maps by generating motion fields via optical flow and then moving the precipitation systems according to these fields, thereby obeying the 2D continuity equation. This physical information is fused into the learned features of NowcastNet's generator via Spatially-Adaptive Denormalisation \cite{ParkSPADE} (SPADE), and, by adversarial training, the full model learns to predict both sharp and physically more correct rain images.


\section{Methods} \label{sec:Methods}
\subsection{Proposed model: MAD-SmaAt-GNet}

\begin{figure*}
    \centering
    \includegraphics[width=1\textwidth]{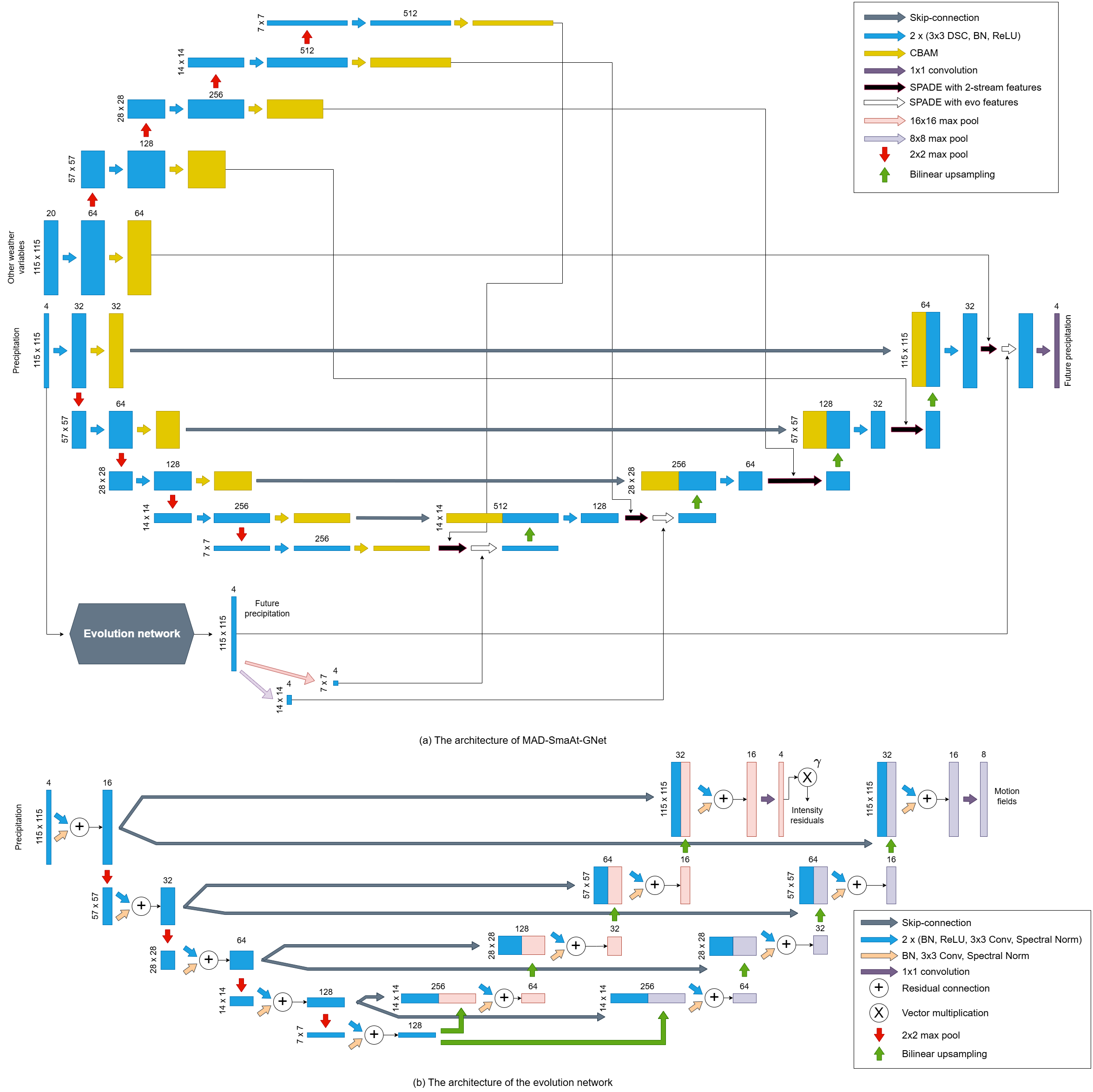}
    \caption{The architecture of (a) MAD-SmaAt-GNet when given $4$ rain images and corresponding images of other weather variables and a target of $4$ output images (best viewed in colour). Additionally, the architecture of (b) the evolution network with the same rain inputs. Each bar represents a sequence of images or feature maps with corresponding dimensions. The number of channels is given above the bars and the height and width of the images on the left of the bars (unless unchanged).}
    \label{fig:Model}
\end{figure*}


The Multimodal Advection-Guided Small Attention GNet model (MAD-SmaAt-GNet) developed in this study extends the SmaAt-UNet architecture \cite{TrebingSmaAtUNet}. It introduces an additional encoder, i.e. an extra stream, to process the other weather variables (multimodal input), giving the architecture a G-Net shape similar to \cite{REULEN2024112612}. In addition, MAD-SmaAt-GNet incorporates the evolution network from NowcastNet \cite{Zhang2023526}, which is an advection-based U-Net \cite{RonnebergerUnet} featuring two decoders and shared skip connections. A schematic overview of the architecture is provided in \autoref{fig:Model}.

As in SmaAt-UNet \cite{TrebingSmaAtUNet}, each encoder and the decoder consist of four levels, i.e., four downsampling and four upsampling operations.
Each downsampling operation of an encoder consists of a max-pooling layer with reduction factor $2$, a double convolutional layer with DSCs \cite{CholletDSC}, and a CBAM \cite{WooCBAM} at the end. Each upsampling operation of the decoder consists of an upsampling layer of bilinear interpolation with scale factor $2$, a double convolutional layer with DSCs \cite{CholletDSC}, and a SPADE \cite{ParkSPADE} layer at the end that either applies dynamic conditioning with both the features of the other weather variables and the features from the evolution network or solely with those of the other weather variables. At the bottleneck, the SPADE \cite{ParkSPADE} is applied before the first upsampling operation. After the final upsampling operation, the output layer, which is a convolutional layer with $1\times1$ kernel, maps the learned representations to the final predictions.

When the model receives its inputs (four rain images and four images each for temperature at 300 m, air pressure, relative humidity, and the U- and V-wind components at 300 m) the rain images are first passed through the evolution network \cite{Zhang2023526}. This evolution network is a U-Net \cite{RonnebergerUnet} with four downsampling and four upsampling stages. The rain input is first projected to 16 channels using a double convolutional layer. Each double convolutional block in the evolution network consists of 2D batch normalisation, a ReLU activation, and a convolution with spectral normalisation \cite{MiyatoSpectralNorm}, applied twice in sequence. A residual shortcut is also included: it applies 2D batch normalisation followed by a single convolution with spectral normalisation \cite{MiyatoSpectralNorm}, and its output is added to the output of the double convolutional block.

After the initial channel expansion, four downsampling operations follow, each halving the spatial resolution and doubling the number of channels. Each down block consists of 2D max pooling followed by a double convolutional layer. Subsequently, four upsampling operations are performed for each of the two decoders: one that predicts vertical and horizontal motion fields and one that predicts intensity residuals. Each upsampling block uses bilinear interpolation followed by a double convolution. Both decoders end with a $1 \times 1$ convolution; additionally, the intensity residuals decoder multiplies its output by a learnable vector $\gamma$, which scales each channel independently. The motion fields decoder outputs the vertical and horizontal motion fields corresponding to the target sequence.

The motion fields and intensity residuals are then passed to the evolution operator \cite{Zhang2023526}, which advects the most recent input rain image using the predicted vertical and horizontal motion field and adds the corresponding intensity residual. This autoregressive procedure is repeated using each newly predicted frame until all target frames have been produced. The final outputs of the evolution network \cite{Zhang2023526} are used in two ways within MAD-SmaAt-GNet: (i) they are fed directly into the model’s final decoder, and (ii) they are passed through two max-pooling layers with reduction factors 16 and 8, enabling the evolution network’s features to be integrated at the bottleneck level and at the first decoder level.

The rain images are also given to the rain encoder which is identical to the encoder of SmaAt-UNet \cite{TrebingSmaAtUNet}, except for the channels being half of those in SmaAt-UNet because of the fewer input images. So, the channels are first incremented to $32$ channels and doubled each downsampling operation.

Before the final decoder stage, the inputs of the additional weather variables are processed by their dedicated encoder, which, analogous to the SmaAt-UNet encoder \cite{TrebingSmaAtUNet}, produces feature maps at four resolution levels using DSCs \cite{CholletDSC} and CBAMs \cite{WooCBAM}. Unlike the rain encoder, this encoder uses twice as many channels, reflecting the greater number of input variables, as illustrated in \autoref{fig:Model}. The resulting feature maps are then passed, at each level, to the corresponding SPADE layer \cite{ParkSPADE} in the decoder.

The decoder receives the rain features from the bottleneck and from each skip-connection of the rain encoder, and subsequently enriches them through SPADE conditioning \cite{ParkSPADE} using the evolution network’s features together with the encoder features of the additional weather variables, or, where applicable, using only the latter. SPADE conditioning is applied at every decoder level after the upsampling operation (or before the upsampling operation at the bottleneck). In contrast, features from the evolution network are used for SPADE conditioning only at the bottleneck, after the first upsampling step, and after the final upsampling step that leads back to the original spatial resolution. This restricted use was adopted because incorporating the evolution network’s features at all decoder levels caused the model to become large and unstable during training. After all upsampling operations, the final fused features are passed to the output layer to generate the predicted rain maps.

\subsection{Benchmark model, ablation study, and parameters}

The benchmark model used in this study is the SmaAt-UNet \cite{TrebingSmaAtUNet}, on which MAD-SmaAt-GNet is based. In addition, a persistence model (predicting all target frames by repeating the last input frame) is included to assess whether the models exhibit predictive skill beyond this simple baseline.



An ablation study was conducted for MAD-SmaAt-GNet. Three variants were evaluated: a model that adds only the evolution network (\emph{SmaAt-UNet with Evo-Net}); a model that adds only the encoder of additional weather variables (\emph{SmaAt-UNet with 2-stream}); and the evolution network evaluated in isolation (\emph{Evo-Net}). These models were compared against the baseline \emph{SmaAt-UNet} \cite{TrebingSmaAtUNet} and the full MAD-SmaAt-GNet to assess the contribution of each architectural component.


\autoref{tab:Params} provides an overview of all models and their parameter counts. As shown, MAD-SmaAt-GNet is larger than SmaAt-UNet \cite{TrebingSmaAtUNet}, yet remains relatively lightweight compared to many other nowcasting architectures \cite{An2025SurveyPrecip}.

\newcolumntype{L}{>{\centering\arraybackslash}m{2cm}} 

\begin{table}
    \renewcommand{\arraystretch}{1.5} 
    \centering
    \caption{Overview of the models and their model sizes in number of parameters.}
    \begin{tabular}{|L|L|L|}
        \hline
        \textbf{Model} & \textbf{Number of parameters} & \textbf{Relative model size} \\
        \hline
        SmaAt-UNet & $4,110,400$ & $1\times$ \\
        MAD-SmaAt- GNet & $7,453,676$ & $1.81\times$ \\
        SmaAt-UNet with Evo-Net & $3,745,936$ & $0.91\times$ \\
        SmaAt-UNet with 2-stream & $4,766,624$ & $1.16\times$ \\
        Evo-Net & $2,219,500$ & $0.54\times$ \\
        Persistence & $-$ & $-$\\
        \hline
    \end{tabular}
    \label{tab:Params}
\end{table}


Due to the much smaller size of the rain inputs compared to the rain inputs in \cite{TrebingSmaAtUNet}, the number of channels of the convolutional layers of certain encoders and decoders were halved. This was done in the rain encoder and decoder for the full model and for each variant model, and for the encoder and two decoders of the evolution network \cite{Zhang2023526}. The reason for this was to keep the models light-weight and to prevent them from overfitting. The number of channels in the convolutional layers of the SmaAt-UNet \cite{TrebingSmaAtUNet} model was kept the same as in \cite{TrebingSmaAtUNet}.

\subsection{Training}

All models were trained using an early-stopping criterion that halted training if the validation loss did not improve for 15 consecutive epochs relative to the best value achieved so far. The mean squared error (MSE) was used as the loss function, matching the choice in SmaAt-UNet \cite{TrebingSmaAtUNet} to ensure a fair comparison.

Similar to \cite{TrebingSmaAtUNet}, the Adam optimiser \cite{KingmaAdamOpt} was used with an initial learning rate of $0.001$ and a learning rate scheduler that reduces the learning rate with a reduction factor of $0.1$ if the validation score did not improve after $5$ epochs (in contrast to \cite{TrebingSmaAtUNet} where a patience of $10$ epochs was used). A batch size of $16$ samples was used when training the models and the models were trained on computer with a NVIDIA GeForce GTX 1050 Ti with 4 GB VRAM.

The models with the evolution network as component (MAD-SmaAt-GNet and SmaAt-UNet with Evo-Net) were trained with and without a pre-trained evolution network, and with and without freezing the pre-trained evolution network's parameters when training the whole model. For both MAD-SmaAt-GNet and SmaAt-UNet with Evo-Net, the best results were obtained when the models were given a pre-trained evolution network but the evolution network's parameters not being frozen during the training of the whole model, thereby allowing these parameters to be further calibrated to work in tandem with the rest of the model.

\subsection{Quality metrics and model evaluation} \label{subsec:Quality}

For model evaluation, the mean squared error (MSE) was also used as the quality metric, consistent with the loss function employed during training. The calculation of the total MSE is shown below:
\begin{equation}
MSE = \frac{1}{K}\frac{1}{M}\frac{1}{N}\sum_{k=0}^K\sum_{j=0}^M\sum_{i=0}^N{(\hat{y}_{ijk}-y_{ijk})^2},
\end{equation}
where $\hat{y}_{ijk}$ and $y_{ijk}$ are the predicted image and ground-truth image respectively for batch $k$, sample $j$, and time step $i$, and $K$ is the number of batches, $M$ the number of samples per batch, and $N$ the number of target images.
Moreover, the predicted rain images were turned into a binary classification by applying a threshold to the pixel values of each image and so creating binary masks. This was done for both the predicted images and the corresponding ground-truth images. To ensure comparability with \cite{TrebingSmaAtUNet}, a threshold of $0.5,\mathrm{mm/h}$ was used.

With the aforementioned binary masks the true positives (TN) and negatives (TN) and false positives (FP) and negatives (FP) can be computed over the whole test set for each threshold. This is done by comparing each pixel of the prediction mask with the ground-truth mask. For example, a prediction of $1$ for a pixel in the prediction mask, i.e. that pixel value is greater than the threshold, with an equivalent ground-truth label of $1$ corresponds to a true positive, and a prediction of $1$ with a ground truth of $0$ to a false positive, etcetera. With those numbers the Accuracy (ACC), Precision (PREC), Recall (REC), F1-score (F1), Critical Success Index (CSI), and Matthews Correlation Coefficient (MCC) can be calculated for the whole test set and for each threshold. The equations for these classification metrics are given below:
\begin{equation}
    ACC=\frac{TP+TN}{TP+TN+FP+FN},    
\end{equation}
\begin{equation}
    PREC=\frac{TP}{TP+FP},
\end{equation}
\begin{equation}
    REC=\frac{TP}{TP+FN},
\end{equation}
\begin{equation}
    F1=2\times\frac{PREC\times REC}{PREC+REC},
\end{equation}
\begin{equation}
    CSI=\frac{TP}{TP+FN+FP},
\end{equation}
\begin{multline}
    MCC= \\
     \frac{TP\times TN-FP\times FN}{\sqrt{(TP+FP)(TP+FN)(TN+FP)(TN+FN)}}.   
\end{multline}
These classification metrics give a good indication of how well the models can actually predict the presence of light to heavy rain. The F1-score and MCC are less prone to imbalances in the classes (positive or negative).

\section{Experiments} \label{sec:Exper}

The dataset that was used to train and evaluate the models was provided by the Royal Dutch Meteorological Institute (KNMI; Koninklijk Nederlands Meteorologisch Instituut). It contained simulations of their numerical HARMONIE model\footnote{More information on the HARMONIE model is available at: \url{https://english.knmidata.nl/open-data/harmonie}} for several variables, and each simulation by the model was done every 6 hours for 48 hours ahead at an hourly interval. The dataset ranged from 2019 to 2023 and consisted of images for each variable where each pixel corresponds to the value of that variable for an area of approximately $2.5$ square kilometres ($0.023^{\circ}$ latitude and $0.037^{\circ}$ longitude per pixel). 

The images were cropped to the geolocation of the Netherlands bounded by the coordinates $[50.84, 53.462]^{\circ}$ N latitude, $[3.182, 7.4]^{\circ}$ E longitude, resulting in a image size of $115 \times 115$. From this dataset, the following variables were extracted: accumulated rain at ground level in kilogrammes per square metres ($kg/m^2$) from the start of the simulation up to any hour within the 48 hours of forecasts; temperature at 300 metres above ground in Kelvin ($K$); air pressure at mean sea level in Pascal ($Pa$); relative humidity at 2 metres above ground in percentage ($\%$); and U- and V-wind component at 300 metres above ground in metres per second ($m/s$). 

\begin{figure*}[h]
    \centering
    \includegraphics[width=0.7\textwidth]{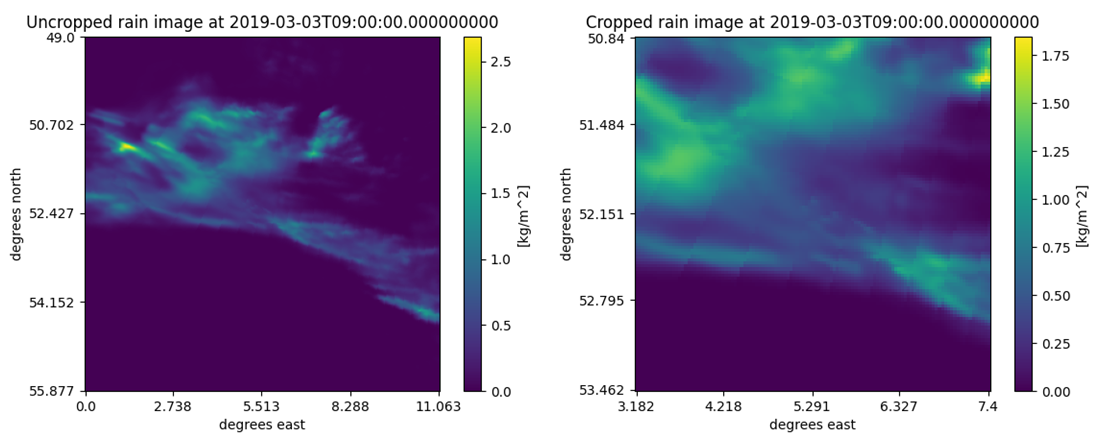}
    \caption{Plots of a rain image before and after the cropping. Note that the coordinates start in the top-left corner and that the images are vertically flipped compared to a standard map of the Netherlands.}
    \label{fig:RainCropping}
\end{figure*}

\begin{figure*}[h]
    \centering
    \includegraphics[width=1\textwidth]{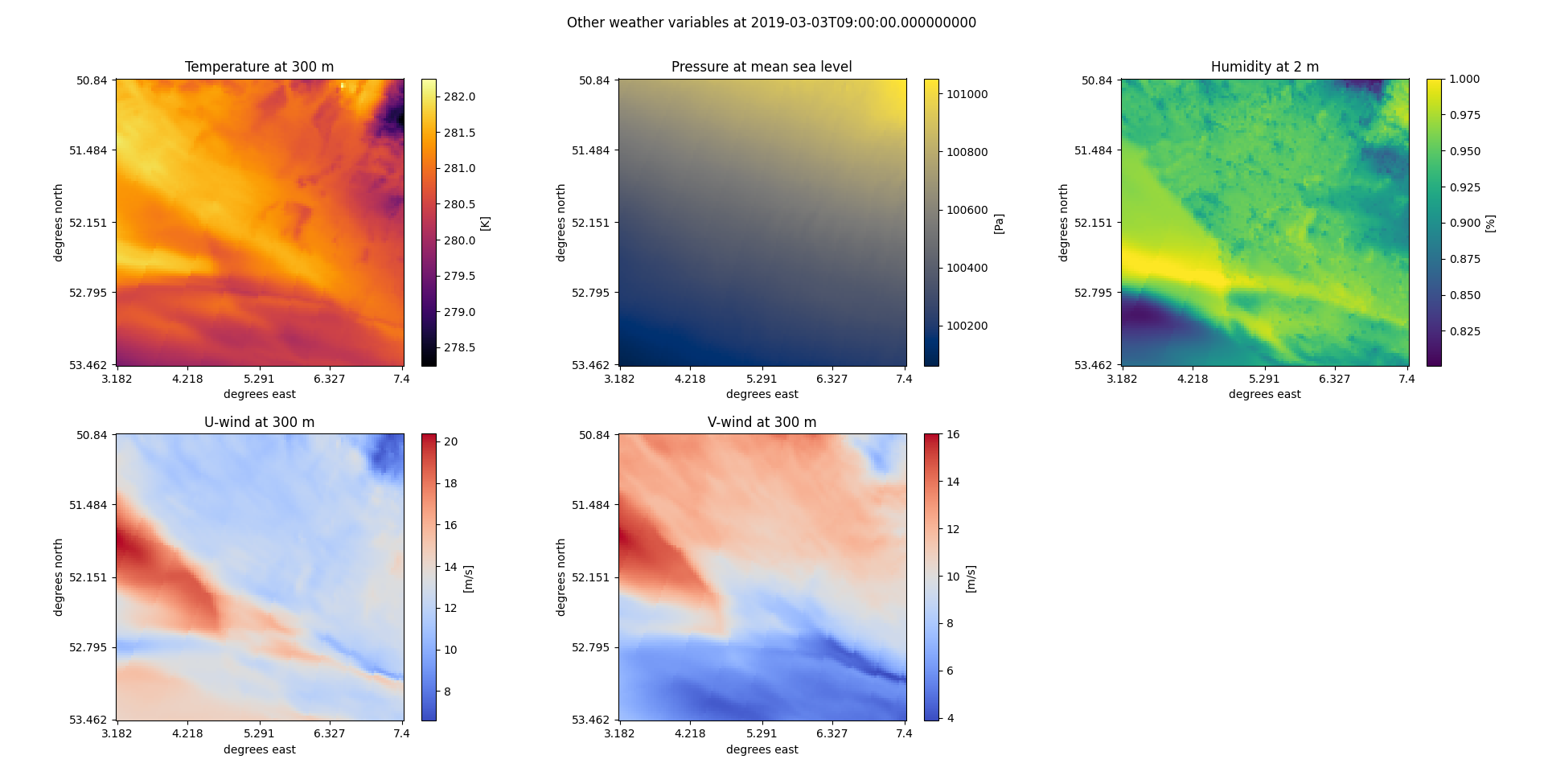}
    \caption{Plots of the other weather variables used as inputs for the models with two streams. Note that the coordinates start in the top-left corner and that the images appear vertically flipped (this is best seen in the image of relative humidity where the map of the Netherlands is somewhat visible from the image).}
    \label{fig:WeatherVars}
\end{figure*}

In \autoref{fig:RainCropping}, a rain image is shown and the cropping of the images is shown for this image with the corresponding coordinates on the axes. The weather variables besides rain that were used are shown in \autoref{fig:WeatherVars}. 

Since there was overlap between simulations as each simulation predicted variables 48 hours ahead but a new simulation also started after each 6 hours, the most recent simulations were used for creating the final dataset. As such, $42,294$ images were obtained for each of the variables. Additionally, the accumulated rain images from the HARMONIE model were turned into rainfall images by subtracting the previous accumulated rain image from the current accumulated rain image, thereby obtaining the rain images in millimetres per hour ($mm/h$), i.e. the rain accumulated over 1 hour (assuming a water density of $1000$ $kg / m^3$). To prevent having many samples with little to no rain and to prevent the models from learning to predict little rain, the samples were only added to the dataset for the models if the first input rain image had $20\%$ of its pixels with a rainfall value of more than $0.1$ $mm/h$. A sample consists of $4$ consecutive input rain images and $4$ consecutive target rain images and the corresponding images for the other weather variables at the same time stamps, $20$ images total for the input of the other weather variables.

As a result, $5,925$ training samples and $1,883$ test samples were created, with samples from 2023 used as the test set. In \autoref{tab:Dataset}, the filtered dataset is shown in comparison to applying no filter. The inputs for the models were normalised beforehand with the corresponding maximum or minimal value for each variable found in the training set, except for relative humidity, which was already expressed as a percentage ranging from $1$ to $0$. The models were subsequently trained with the processed rain images from the filtered dataset and these rain images were also set as the ground-truth images. The models were tasked to predict four future rain images at 1, 2, 3, and 4 hours ahead. Additionally, the MSE was also calculated per predicted image (i.e. per time step) to see how the predictions fare with the increasing time horizon.

\newcolumntype{D}{>{\centering\arraybackslash}m{1.35cm}} 

\begin{table}
    \centering
    \caption{Size of the filtered dataset compared to the whole dataset.}
    \begin{tabular}{|c|D|D|c|}
        \hline
        \multicolumn{4}{|c|}{\textbf{Dataset}} \\
        \hline
        \textbf{Filter} & \textbf{Number of training samples} & \textbf{Number of test samples} & \textbf{\%} \\
        \hline
        No filter & $33,743$ & $8,270$ & $100\%$ \\
        $20\%>0.1$ $mm/h$ & $5,925$ & $1,883$ & $18.6\%$ \\
        \hline
    \end{tabular}
    \label{tab:Dataset}
\end{table}

\section{Results and discussion} \label{sec:ResuDiscu}
\begin{table*}
    \centering
    \caption{
     Test results for the models on the task of predicting a sequence of four consecutive rain images. All classification metrics were computed using the threshold of $0.5$ $mm/h$. Metrics where lower values indicate better performance are marked with a $\downarrow$ symbol, while those where higher is better are marked with a $\uparrow$ symbol. The best results are shown in bold, and the second-best results are underlined.}
    \begin{tabular}{|c|c|c|c|c|c|c|c|}
        \hline
        \textbf{Model} & \textbf{MSE$\downarrow$} & \textbf{Accuracy$\uparrow$} & \textbf{Precision$\uparrow$} & \textbf{Recall$\uparrow$} & \textbf{F1$\uparrow$} & \textbf{CSI$\uparrow$} & \textbf{MCC$\uparrow$} \\
        \hline
        Persistence & $1.0625$ & $0.7989$ & $0.4271$ & $0.5248$ & $0.4709$ & $0.3080$ & $0.3512$ \\ 
        SmaAt-UNet & $0.4721$ & $0.8586$ & $0.5932$ & $0.5440$ & $0.5675$ & $0.3962$ & $0.4839$ \\
        MAD-SmaAt-GNet & $\mathbf{0.4299}$ & $\mathbf{0.8672}$ & $\mathbf{0.6123}$ & $\underline{0.6033}$ & $\underline{0.6078}$ & $\underline{0.4366}$ & $\mathbf{0.5279}$ \\
        SmaAt-UNet with Evo-Net & $\underline{0.4352}$ & $\underline{0.8652}$ & $\underline{0.6034}$ & $\mathbf{0.6125}$ & $\mathbf{0.6079}$ & $\mathbf{0.4367}$ & $\underline{0.5266}$ \\
        SmaAt-UNet with 2-stream & $0.4599$ & $0.8591$ & $0.5861$ & $0.5914$ & $0.5888$ & $0.4172$ & $0.5038$ \\
        Evo-Net & $0.4884$ & $0.8581$ & $0.5835$ & $0.5874$ & $0.5854$ & $0.4139$ & $0.4998$ \\
        \hline
    \end{tabular}
    \label{tab:ResultsSeq}
\end{table*}

\begin{figure}[h]
    \centering
    \includegraphics[width=0.40\textwidth]{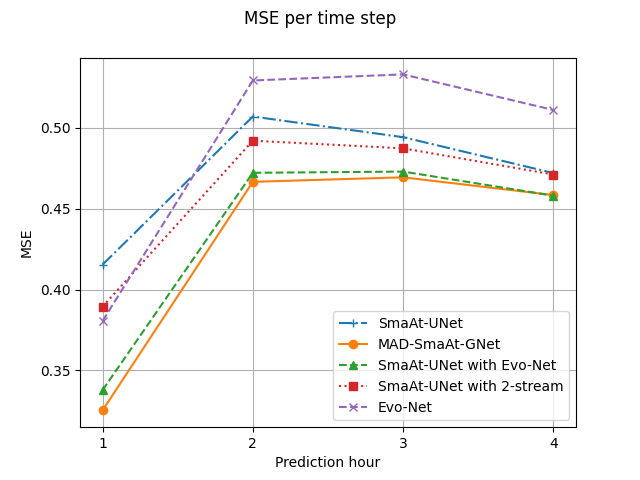}
    \caption{Plot of the MSEs of the models per time step. The persistence model is not included in the plot due to its much higher MSEs which would skew the plot.}
    \label{fig:MSEplot}
\end{figure}

As shown in \autoref{tab:ResultsSeq}, MAD-SmaAt-GNet achieves the best overall performance across nearly all metrics when evaluating the full sequence of predicted images, delivering an 8.9\% reduction in MSE compared with SmaAt-UNet \cite{TrebingSmaAtUNet}. The next best performance is that of the SmaAt-UNet enhanced with the evolution network, followed by the variant that incorporates an additional encoder for additional weather variables. These results demonstrate that every extension of the base model SmaAt-UNet \cite{TrebingSmaAtUNet} yields improved predictions over the original model. Furthermore, \autoref{tab:ResultsSeq} and \autoref{fig:MSEplot} show that all proposed models outperform the persistence baseline across all metrics.

\begin{figure*}
    \centering
    \includegraphics[width=1\textwidth]{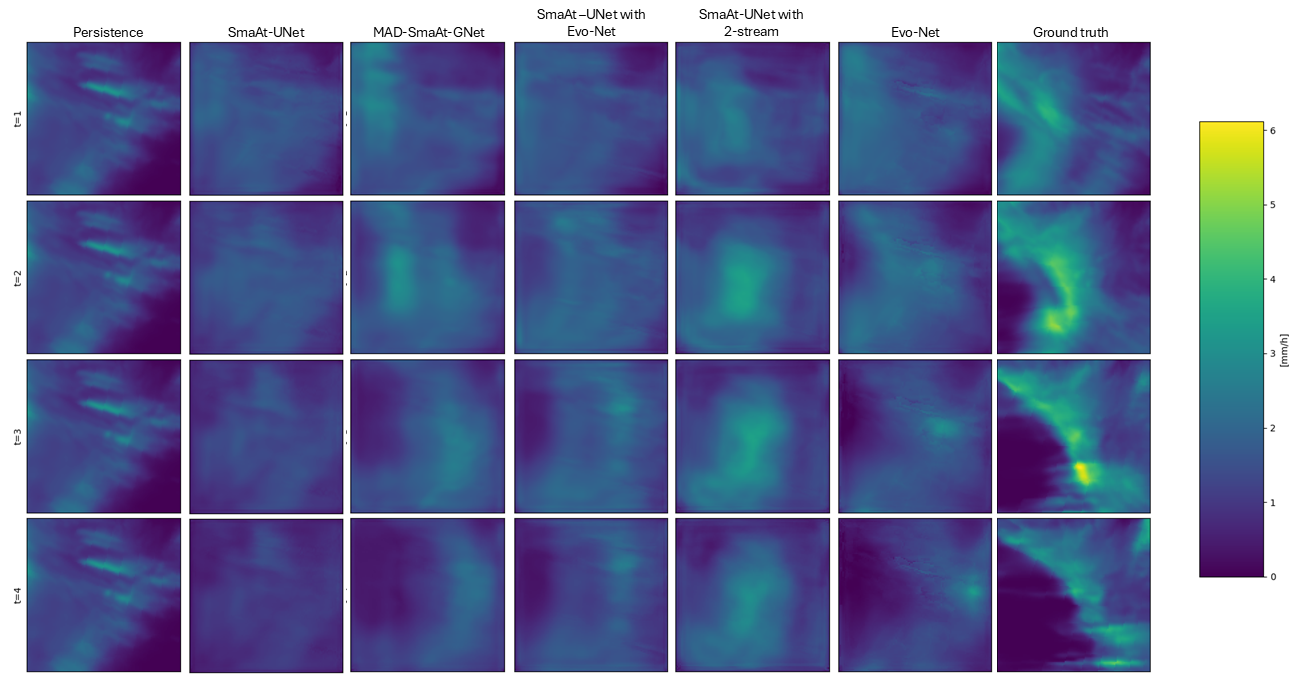}
    \caption{The predictions of the models for a certain test sample and the corresponding ground-truth images. The images form a sequence of predictions from $1$ to $4$ hours ahead.}
    \label{fig:ModelPreds}
\end{figure*}

When examining the performance per forecast lead time, i.e., per predicted image, in \autoref{fig:MSEplot}, it becomes clear that the benefit of incorporating other weather variables (SmaAt-UNet with 2-stream) is most pronounced in the short term, between 1 and 3 hours ahead, but fades by the fourth hour. In contrast, the integration of the evolution network \cite{Zhang2023526} consistently enhances predictions across the entire sequence, as illustrated in \autoref{fig:ModelPreds}. Notably, the SmaAt-UNet with Evo-Net even achieves a slightly lower MSE than MAD-SmaAt-GNet at the final prediction hour. This suggests that while additional weather inputs provide valuable contextual information for rainfall prediction, their usefulness diminishes with longer lead times (at the fourth hour, the MSE is slightly higher than that of the SmaAt-UNet \cite{TrebingSmaAtUNet}). Meanwhile, embedding physical principles delivers sustained improvements, even at longer horizons. Together, these findings imply that extra atmospheric variables can meaningfully boost short-term forecasting, while physically grounded methods benefit both short- and long-range predictions.

\autoref{fig:MSEplot} also shows that the MSE of the models generally increases with longer lead times but dips slightly at hour four. This is likely because precipitation often dissipates by that time, reducing pixel intensities and, consequently, the MSE. The figure omits the persistence baseline as including its MSE values would skew the plot due to its substantially larger MSE. Furthermore, it is apparent that MAD-SmaAt-GNet leads at every time step with SmaAt-UNet with Evo-Net trailing closely behind, except for the last hour where SmaAt-UNet with Evo-Net has a slightly lower MSE.

\autoref{fig:ModelPreds} gives an example of the predicted rainfall fields from each model alongside the ground-truth images for all forecast steps. As noted in \cite{TrebingSmaAtUNet}, the predictions appear blurry because the models are trained using MSE, which encourages predicting average pixel intensities and therefore smooths sharp spatial features. Despite this effect, the models still exhibit distinguishable strengths. The SmaAt-UNet with Evo-Net (and the evolution network \cite{Zhang2023526} alone) more accurately captures the movement of the precipitation system. In contrast, the SmaAt-UNet with 2-stream better matches the intensity of rainfall, which is often underestimated in other models.

MAD-SmaAt-GNet appears to strike a balance between these strengths: it better reproduces the motion of the rain system, similar to the Evo-Net variant, while also generating higher (and thereby more realistic) rain rates, similar to the 2-stream variant. This combination suggests that MAD-SmaAt-GNet leverages the benefits of both extensions. For example, in \autoref{fig:ModelPreds}, MAD-SmaAt-GNet forecasts a larger rain-free area on the left side at 3 and 4 hours ahead, closer to the ground truth than the Evo-Net model, though still not perfectly shaped. It also predicts a strip of heavier rainfall at 2, 3, and 4 hours ahead that corresponds better to the observed field. Overall, MAD-SmaAt-GNet provides the most faithful representation of the evolving precipitation system among the models considered.

Taken together, the results demonstrate that adding additional weather variables and incorporating a physically grounded component both yield measurable improvements individually in SmaAt-UNet with 2-stream and SmaAt-UNet with Evo-Net, and even more when combined in MAD-SmaAt-GNet. \autoref{fig:MSEplot} further suggests that these benefits take effect over different forecast horizons. For practical nowcasting, typically up to about 2 hours ahead \cite{NIU2025M4Caster, niu2024fsrgan, ravuri2021skilful}, both enhancements are meaningfully beneficial. For longer lead times beyond 4 hours, however, one might reasonably omit the additional input variables for simplicity and retain only the physically consistent component.

\section{Conclusions} \label{sec:Conclu}


This paper introduced the MAD-SmaAt-GNet model, which combines information from both a physically consistent component and additional atmospheric variables. The results show that this integrated approach improves short-term precipitation forecasts, which is central to the nowcasting task, outperforming the models that apply only one of these enhancements. The experiments further indicate that physically informed components continue to benefit predictions at longer lead times, whereas the advantage of incorporating additional weather variables diminishes over time. 

\section{Acknowledgements} \label{sec:Acknow}
We would like to acknowledge L. Phaf and S. Tijm from KNMI who manually obtained the weather simulation data for us and did this free of charge for the sake of this study.

\section{Code and data availability} \label{sec:Code}
The implementation of our models is available at GitHub \url{https://github.com/Rogue-Juan/MAD-SmaAt-GNet}. The dataset is available upon request. Please contact \texttt{s.mehrkanoon@uu.nl} if interested.
\bibliography{related_works} \label{sec:Biblio}

\end{document}